\documentclass{article}
\usepackage{spconf,amsmath,graphicx}
\usepackage{amsmath,amssymb,amsfonts}
\usepackage{algorithmic}
\usepackage{graphicx}
\usepackage[numbers,sort&compress]{natbib}
\usepackage{diagbox}
\usepackage{multirow}
\usepackage{multicol}
\usepackage{booktabs}

\usepackage[table,xcdraw]{xcolor}
\title{SX-Stitch: An Efficient VMS-UNet Based Framework for Intraoperative Scoliosis X-Ray Image Stitching}
\name{Yi Li$^1$, Heting Gao$^1$, Mingde He$^1$, Jinqian Liang$^{2,\star}$, Jason Gu$^{3}$,Wei Liu$^{1,\star}$\thanks{$^\star$ Corresponding Author.} \thanks{$^\dagger$ This research was supported by the 2023 Guangdong Basic and Applied Basic Research Fund Regional Joint Fund - Key Project (Project No.2023B1515120017),the 2023 Key Project of Guangdong Provincial Department of Education for General Universities (Project No.2023ZDZX3024) and the National Natural Science Foundation of China (Project No.82072477).}}

\address{$ ^1$Department of Mechanical and Energy Engineering, Southern University of Science and Technology\\$ ^2$Peking Union Medical College Hospital\\$ ^3$Electrical and Computer Engineering, Dalhousie University}

\begin{document}
%\ninept
%
\maketitle
\begin{abstract}
In scoliosis surgery, the limited field of view of the C-arm X-ray machine restricts the surgeons' holistic analysis of spinal structures. This paper presents an end-to-end efficient and robust intraoperative X-ray image stitching method for scoliosis surgery,named SX-Stitch. The method is divided into two
stages:segmentation and stitching. In the segmentation stage, we propose a medical image segmentation model named Vision Mamba of Spine-UNet (VMS-UNet), which utilizes the state space Mamba to capture long-distance contextual information while maintaining linear computational complexity, and incorporates the SimAM attention mechanism, significantly improving the segmentation performance.In the stitching stage, we simplify the alignment process between images to the minimization of a registration energy function. The total energy function is then optimized to order unordered images, and a hybrid energy function is introduced to optimize the best seam, effectively eliminating parallax artifacts. On the clinical dataset, Sx-Stitch demonstrates superiority over SOTA schemes both qualitatively and quantitatively.

\end{abstract}
\section{Introduction}
\label{sec:intro}
Intraoperative spinal images are generally obtained through the use of small to medium-sized, mobile C-arm X-ray machines. However, constrained by their limited field of view (FOV), surgeons can only acquire truncated slices of the spinal image. The process of combining these truncated image slices into a panoramic view is known as image stitching.

\begin{figure}[t]\centering
	\includegraphics[width=8.5cm]{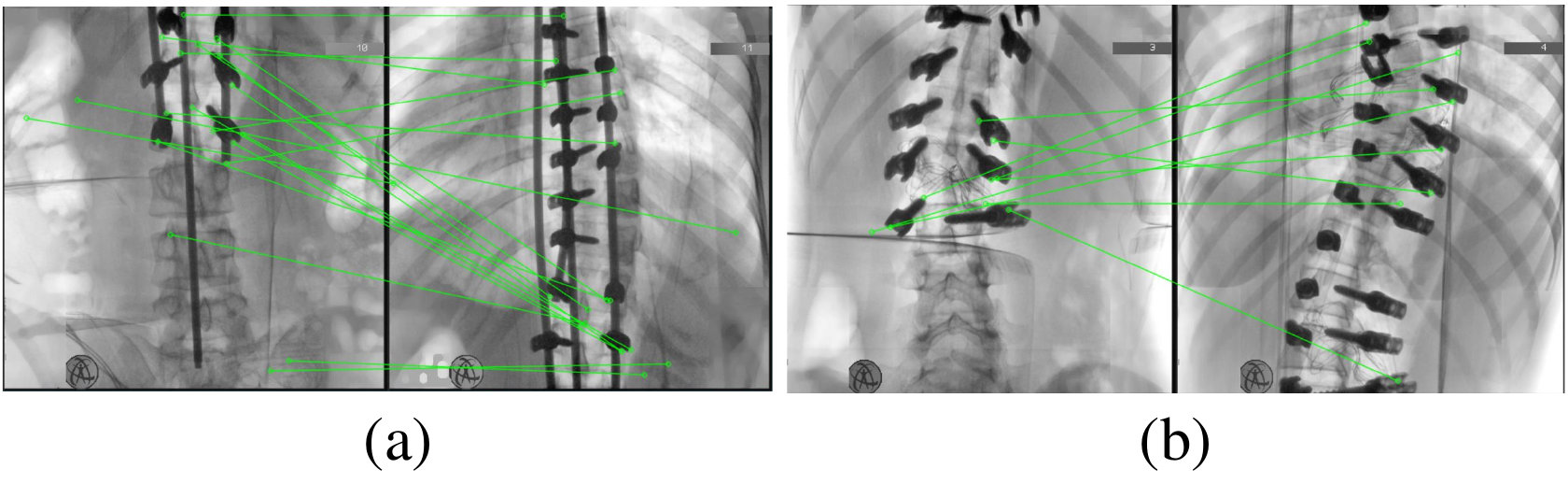}
	\caption{SIFT matching on X-ray image.Due to weak features and repetitive textures of X-ray images, manually designed features perform poorly in robustness.}
    \label{SIFT}
\end{figure}

 In image stitching, the homography transformation is a commonly used model for warping images, which includes translation, rotation, scaling, and viewpoint transformation, accurately explaining the transformation from one two-dimensional plane to another.
 
In image stitching of natural scenes,traditional image stitching methods utilize manually defined feature detection techniques to calculate the homography matrix\cite{AutoStitch,ELA,jia2021leveraging,feng2024seam,multi}. The core idea of these methods is to design optimal features (points, lines, or energy functions) to achieve image alignment.However, the spinal X-ray images captured by C-arms, due to their weak features and repetitive textures, manually designed features exhibit low robustness because of their complex feature design, as shown in Figure  \ref{SIFT}. To achieve more robust and generalizable stitching, researchers have proposed deep learning-based methods to calculate the homography matrix\cite{detone2016deep,nie2021unsupervised,nie2023parallax}.bypasses feature detection and instead directly use Convolutional Neural Networks (CNNs) to extract feature information from matched image pairs and input it into a regression model to estimate the homography matrix in a parameterized manner.By training on a large number of image pairs and optimizing through backpropagation, optimal alignment can be achieved. However, deep learning methods require tens of thousands of image pairs for training (UDIS\cite{nie2021unsupervised} and UDIS2\cite{nie2023parallax}), which is unfeasible for intraoperative truncated spinal X-ray images due to the insufficient amount of data. This limitation can make it difficult to converge to an effective, generalized stitching model. Consequently, these stitching methods, effective for natural scenes, cannot yet be directly applied to X-ray image stitching.

\begin{figure}[t]\centering
	\includegraphics[width=9cm]{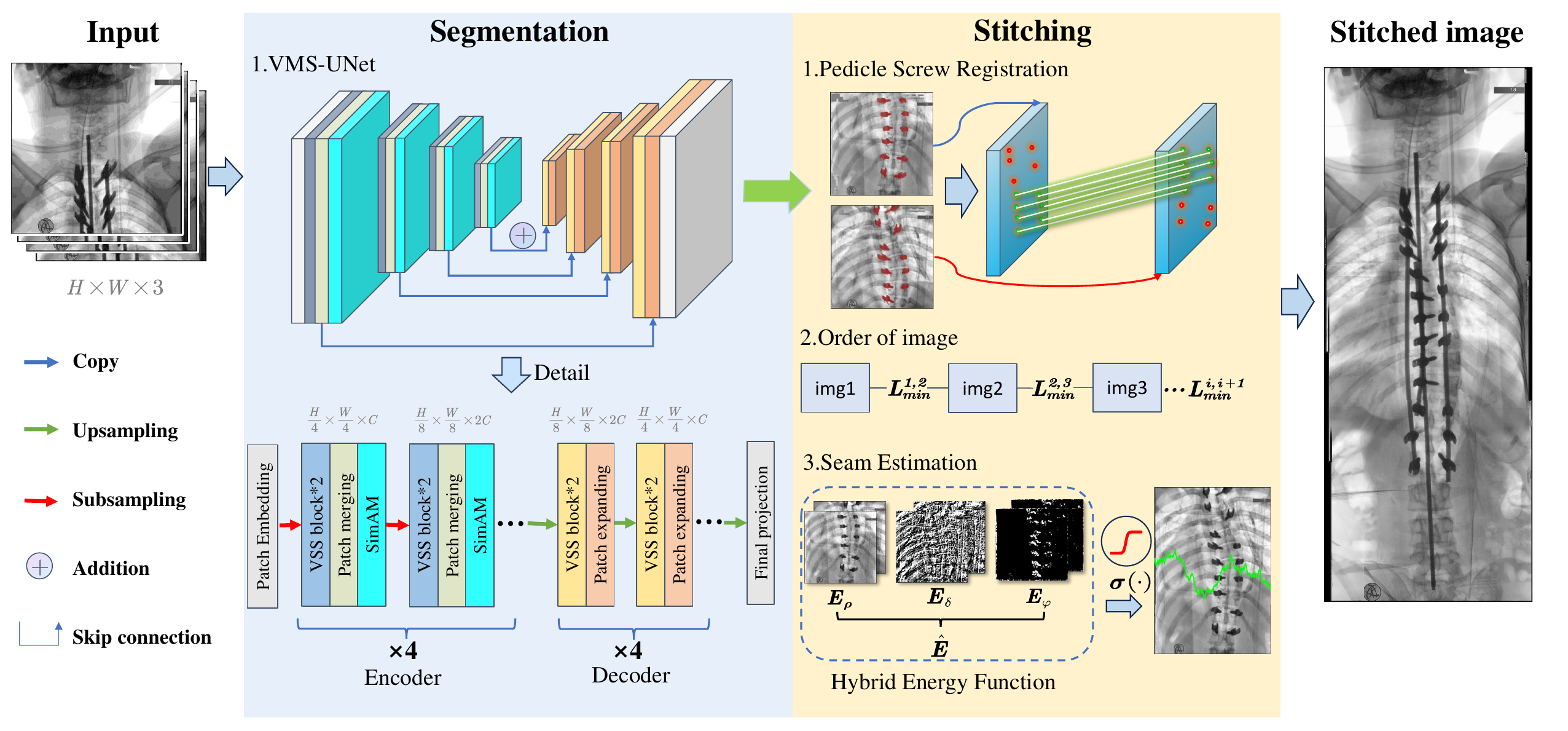}
	\caption{Pipline of SX-Stitch.}
    \label{Pipline}
\end{figure}

Therefore, for the task of X-ray image stitching in non-natural scenes, researchers have proposed various image stitching methods, including those based on additional markers or structures\cite{marker2,marker3}, local image features\cite{feature1,feature2}, and pixel-based approaches \cite{pixel1,pixel2,pixel3}.Methods that rely on additional markers or structures tend to be more complex to operate and may lead to increased costs. In contrast, methods that depend on local features and pixels exhibit insufficient robustness.
In addition, Fotouhi et al. \cite{deep-reconstrcut} proposed an end-to-end long bone image stitching approach that uses CNNs for 2D reconstruction of multiple images. The method employs SSIM and adversarial loss to enforce the network to generate images that are visually similar to the ground truth. However, this method is currently designed only for the femur and has not yet been extended to other types of bones. To address the above challenges, this paper introduces a rapid and robust end-to-end method for stitching full-length spinal images for scoliosis,named SX-stich. The pipeline is depicted in the Figure \ref{Pipline}.

Inspired by the success of VisionMamba\cite{mamba} in image classification tasks and VM-Unet\cite{vm} in medical image segmentation, we have introduced an improved version of the Mamba model-based VM-UNet network, named Vision
Mamba of Spine-UNet(VMS-UNet). This network enhances semantic information perception while maintaining linear complexity, and it performs well on images with sparse features.Subsequently, image registration is performed based on the segmented pedicle screws. In scoliosis surgery, fixing the corrected spine with pedicle screws is an important step, and there is a clear correspondence between the screws. Therefore, we introduce a specific registration energy function that minimizes the distance between corresponding screws to achieve alignment.In multi-image stitching, the output of the energy function guides the image sorting.Ultimately, to eliminate seams and artifacts in the stitched image, we designed a hybrid energy function from three aspects: image pixel, geometric structure, and semantic features, to estimate the optimal seam.

\section{Methodology}
\subsection{VMS-UNet}

 To achieve precise segmentation of the pedicle screw area, we constructed a neural network to perform the semantic segmentation task.UNet\cite{Unet}, as one of the models based on CNNs, is widely praised for its simple structure and strong scalability. However, the standard UNet, due to its limited receptive field, mainly captures local features and struggles to extract information from the global image.  Ruan et al.\cite{vm} proposed a novel UNet architecture based on a state space model—VM-Unet.VM-Unet is not only capable of capturing a wide range of contextual information but also maintains linear computational complexity, providing an efficient solution for medical image segmentation.
We adopted the VM-UNet network architecture and made improvements by integrating a lightweight attention module: SimAM(A Simple, Parameter-Free Attention Module for Convolutional Neural Networks)\cite{simam}. Specifically, spinal X-ray images (H×W×3) are input into an asymmetric encoder-decoder structure. During the encoding and decoding phases, Patch merge and Patch expanding perform downsampling and upsampling, mapping the channels to [8C, 4C, 2C, C], respectively. The VSS block is the core module of VM-UNet, capable of capturing extensive contextual information.

The information after downsampling is not directly passed to the decoder phase. Instead, it is passed through the lightweight parameter-free attention module SimAM to further focus on the key features of the model. SimAM can generate three-dimensional attention weights by computing the local self-similarity of the feature maps. 

The features processed by SimAM, serving as the down-sampled features, are combined with the features from the decoder phase that have passed through the VSS block via addition to achieve skip connection. The connected features are then input into Patch Expanding for up-sampling.

\subsection{Pedicle screw registration}
Given a pair of input images to be stitched, $ I_1,I_2 $ , after segmentation, we utilize the obtained masks to calculate the geometric centroid $S$ of each pedicle screw in both images.

To ensure precise alignment of the corresponding screws in a pair of images, we apply a homography matrix \( H \) to distort $ I_2 $. In the process of determining the homography matrix, we introduce a registration energy function that optimizes \( H \) to minimize the differences between the two images:

\begin{equation}
\label{Lalign}
\begin{aligned}
L_{align} = \sum_{\left( S_{1}^{i}, S_{2}^{j} \in I_1 \cup H\left( I_2 \right) \right)}^N \hspace{-4ex} \ {min \lVert S_{1}^{i} - H\left( S_{2}^{j} \right) \rVert ^2}
\end{aligned}
\end{equation}

Where $i$ and $j$ denote the indices of the screw centroids in the two images, respectively. \( H(\cdot) \) signifies the homographic transformation, which is applied to distort $ I_2 $. $\cup$ represents the calculation of the overlapping region between the distorted target image $ I_2 $ and the reference image \( I_1 \).

\subsection{From disorder to order}
When the stitching process is extended to multiple images, the primary task is to determine the stitching order between images. Through Equation \ref{Lalign}, we can infer that: when all images are arranged in the correct order, the sum of the registration energy function  \( L_{\text{align}} \) between them will be minimized.Therefore, our optimization goal is to minimize this cumulative sum:
\begin{equation}
\begin{aligned}
E_p=min\sum_{i=1,j\ne i}^N{L_{align}^{}}
\end{aligned}
\end{equation}
In this process, \( i \) and \( j \) represent the indices of the images, respectively. During the optimization, each image is treated as a node in the graph, and the registration function value between image pairs is considered as the distance between nodes. After selecting one image (a node), we choose another image with the minimum registration function value (i.e., the closest distance) as the next node.

After determining the image order, we use the registration function to calculate the homography matrix \( H_{i,j}^R \) between adjacent images, taking the topmost image as the reference image, and the transformations of the other images can be calculated based on the accumulated product of the homographies:$ I_k=\prod_{i=1}^k{H_{R}^{i,j}}\cdot I_1\text{,}j=2\cdots k+1 $ .Where  \( k \) represents the correct order of the images, and \( I_k \) denotes the \( k \)-th image in the sequence.

\subsection{Seam Estimation}
In the image fusion process, we designed a seam path optimization algorithm based on a hybrid energy function.

Color difference energy measures the pixel differences between two images in the grayscale space:
\begin{equation}
E_{\rho}\left( u,v \right) =\left( I_1\left(u,v \right) -I_2\left(u,v \right) \right) ^2
\end{equation}
The geometric structure energy function calculates the gradients in both horizontal and vertical directions:
\begin{equation}
E_{\delta}\left( u,v \right) =\left( \varDelta _1\left( u,v \right) -\varDelta _2\left( u,v \right) \right) ^2
\end{equation}
 in which $ \varDelta(\cdot)$ represent the square sum of the gradients in the x and y directions.

The deep feature difference calculates the feature energy, specifically, we use the 24th layer of a pre-trained ResNet-50\cite{resnet} as the representation of image semantic content to compute the difference:
\begin{equation}
E_{\varphi}\left( u,v \right) =\left( \varPhi _1\left( u,v \right) -\varPhi _2\left( u,v \right) \right) ^2
\end{equation}
In which $ \varPhi(\cdot)$ represents the ResNet50 features.
The hybrid energy is defined as the combination of various energy components:
\begin{equation}
\hat{E}\left( u,v \right) =\lambda _{\rho}E_{\rho}\left( u,v \right) +\lambda _{\delta}E_{\delta}\left( u,v \right) +\lambda _{\varphi}E_{\varphi}\left( u,v \right) 
\end{equation}
In which  $ \lambda $ represents the weight factor for each energy term.

We first initialize the energy at the starting point of the seam, then expand downwards from the starting point. During each expansion, we select pixel with the minimum energy as the growth point. When the expansion reaches the last column, we backtrack along the determined optimal path to establish the final seam path.The fusion image is calculated as:
\begin{equation}
I_{seam}\left[ :,v \right] =\sigma \left( kv \right) \odot I_1\left[ :,v \right] +\left( 1-\sigma \left( kv \right) \right) \odot I_2\left[ :,v \right] 
\end{equation}
where  \( \sigma(\cdot)\) represents the Sigmoid function, and \( k \) serves as an amplification factor, which is used to achieve a progressive fusion effect on both sides of the seam.

\section{Experiment and result}
\subsection{Dataset and Implement Details}
The segmentation part of the training data comprises 1,032 C-arm intraoperative spinal X-ray images provided by clinical hospitals, including images with resolutions of 512×512, 1024×1024, and 1920×1920.The image set encompasses consecutive truncated images all originating from the same patient and having an overlap ranging from 20\% to 90\%. Screws have been implanted in the bones. All model framework components are implemented on the PyTorch platform. Testing and training operations are conducted on a single GPU equipped with an NVIDIA RTX 3070.

\begin{table*}[h]
\normalsize
\centering
\caption{Qualitative comparison of different methods on clinical datasets.
} % 添加标题
\label{table_qualiative} % 设置标签
\resizebox{\linewidth}{!}{
\begin{tabular}{@{}ccccccc|ccccccccc@{}}
\toprule
\multicolumn{1}{l}{\multirow{3}{*}{Algorithm}} & \multicolumn{6}{c|}{Different overlap rates}                                                       & \multicolumn{9}{c}{Different resolutions}                                                                                                          \\ \cmidrule(l){2-16} 
\multicolumn{1}{l}{}                           & \multicolumn{3}{c}{SSIM}                         & \multicolumn{3}{c|}{PSNR}                       & \multicolumn{3}{c}{SSIM}                        & \multicolumn{3}{c}{PSNR}                         & \multicolumn{3}{c}{Elapsed time}              \\
\multicolumn{1}{l}{}                           & 20-40          & 40-70          & 70-90          & 20-40         & 40-70          & 70-90          & 512           & 1024           & 1920           & 512            & 1024           & 1920           & 512           & 1024          & 1920          \\ \midrule
Manually                                       & 0.433          & 0.457          & 0.574          & 18.49         & 18.24          & 19.79          & 0.488         & 0.421          & 0.465          & 18.84          & 17.45.         & 18.01          & ——            & ——            & ——            \\
AutoSitich\cite{AutoStitch}                                     & 0.234          & 0.162          & 0.114          & 11.45         & 13.43          & 14.98          & 0.17          & 0.191          & 0.207          & 13.28          & 12.48          & 13.61          & 25.33         & 30.45         & 42.56         \\
ELA\cite{ELA}                                           & 0.482          & 0.342          & 0.567          & 16.71         & 18.11          & 18.76          & 0.464         & 0.484          & 0.554          & 16.19          & 17.22          & 17.78          & 18.64         & 24.58         & 32.76         \\
LPC\cite{jia2021leveraging}                                            & 0.505          & 0.518          & 0.631          & 15.44         & 18.73          & 19.84          & 0.501         & 0.539          & 0.637          & 17,44          & 17.63          & 20.01          & 12.55         & 17.99         & 27.93         \\
UDIS \cite{nie2021unsupervised}                                          & 0.542          & 0.567          & 0.621          & 16.33         & 18.54          & 19.66          & 0.576         & 0.578          & 0.612          & 18.51          & 19.33          & 19.57          & 4.87          & 6.18          & 15.29         \\
UDIS2\cite{nie2023parallax}                                          & 0.429          & 0.294          & 0.609          & 15.23         & 17.83          & 18.01          & 0.444         & 0.491          & 0.593          & 17.02          & 18.73          & 19.65          & \textbf{3.96} & 4.93          & 12.44         \\
Ours                                           & \textbf{0.633} & \textbf{0.656} & \textbf{0.751} & \textbf{20.5} & \textbf{21.69} & \textbf{23.63} & \textbf{0.68} & \textbf{0.775} & \textbf{0.793} & \textbf{21.94} & \textbf{23.64} & \textbf{25.77} & 4.33          & \textbf{4.76} & \textbf{5.03} \\ \bottomrule
\end{tabular}
}
\end{table*}

\subsection{Quantitative Comparison}

 We compared our approach with traditional feature-based stitching solutions in natural scenes, represented by AutoStitch
 \cite{AutoStitch}, ELA\cite{ELA}, LPC\cite{jia2021leveraging}, as well as the currently popular deep learning stitching frameworks UDIS\cite{nie2021unsupervised} and UDIS2\cite{nie2023parallax}.To compare the effectiveness of our distortion scheme, we invited  clinical doctors to perform manual stitching.

The stitching results across various overlap rates and resolutions are as shown in Table \ref{table_qualiative}. The experimental outcomes demonstrate that our proposed stitching approach excels among all compared methods. Traditional stitching solutions currently provide inferior stitching quality in the majority of cases, sometimes even failing to stitch. Deep learning-based stitching solutions underperform with image pairs of low overlap rates. In contrast, our scheme is adaptable to different overlap rates and resolutions, achieving higher registration quality even under the conditions of minimal overlap and lowest resolution. Furthermore, our method matches the speed of deep learning-based methods but maintains the highest computational efficiency at higher resolutions, with the shortest overall processing time, proving the real-time capability of our approach.

\label{sec:format}
\begin{figure}[t]\centering
	\includegraphics[width=8.5cm]{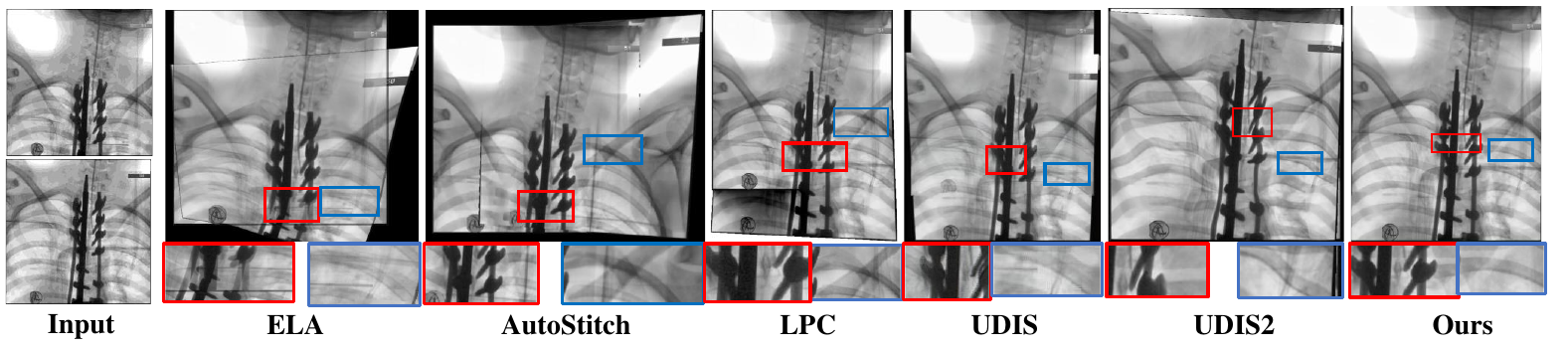}
	\caption{Qualitative Comparison on Paired Image Stitching}
    \label{qualiative_comparison}
\end{figure}

\label{sec:format}
\begin{figure}[t]\centering
	\includegraphics[width=8.5cm]{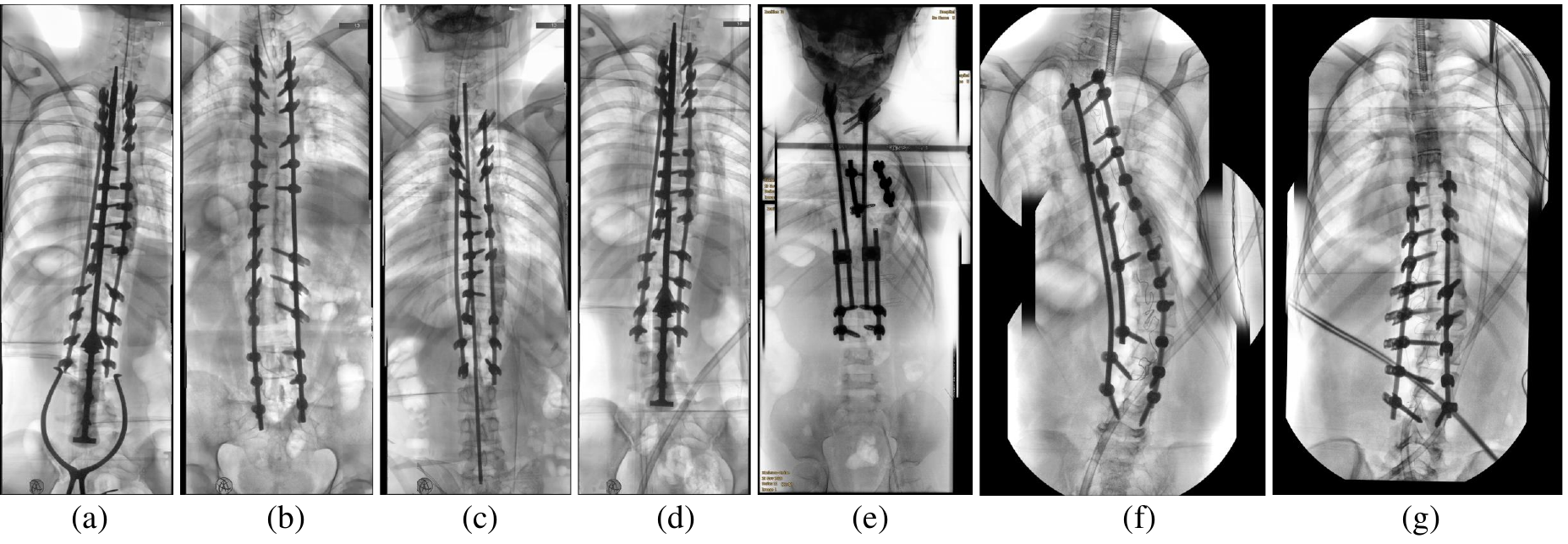}
	\caption{Qualitative Comparison on Multi-Image Stitching.}
    \label{Qualitative Comparison on Multi-Image Stitching}
\end{figure}

\subsection{Qualitative Comparison}

\textbf{Paired Image Stitching.} The qualitative results are displayed in the Figure \ref{qualiative_comparison}, where we first performed stitching on paired images. We paid particular attention to the clarity of bone structures and the correct alignment of pedicle screws, thus deliberately enlarging key detail areas . By analyzing Figure \ref{qualiative_comparison}, we found that ELA\cite{ELA} has inaccuracies in image content alignment, characterized by twisted image distortion and noticeable artifacts. AutoStitch\cite{AutoStitch} also fails to achieve correct image alignment in some cases, while LPC\cite{jia2021leveraging}, although improving alignment accuracy, still has unnatural distortions and shadows. UDIS\cite{nie2021unsupervised} has reduced some artifacts but its alignment accuracy is weak, and UDIS2\cite{nie2023parallax} leads to distortion, especially with abnormal bending of the fixation rod of the pedicle screws. In contrast, our method accurately aligns the image content without introducing artifacts or distortion, providing high-quality stitching effects.

\textbf{Multi-Image Stitching.}  Figure \ref{Qualitative Comparison on Multi-Image Stitching}  clearly demonstrates the high accuracy of our stitching results in content alignment, and the progressive fusion technique applied effectively reduces some seams and parallax artifacts. The structure of the pedicle screws and fixation rods is clearly visible overall, which allows the physiological curvature and force line of the spine to be clearly displayed.

\begin{table}[t]
\centering
\caption{Ablation study on Number of SimAM:Segment and stitch performance} % 添加标题
\label{table_Ablation} % 设置标签
\resizebox{\linewidth}{!}{\begin{tabular}{@{}ccccc@{}}
\toprule
                               & \multicolumn{2}{c}{Segment performance}                                       & \multicolumn{2}{c}{Stitch performance}                                       \\
\multirow{-2}{*}{Num of SimAM} & Acc(\%)                              & mIoU(\%)                              & PSNR                                  & SSIM                                 \\ \midrule
0                              & 85.64                                 & 69.55                                 & 18,74                                 & 0.583                                \\
1                              & 86.85                                 & 70.89                                 & 18.88                                 & 0.59                                 \\
2                              & 88.91                                 & 73.45                                 & 19.82                                 & 0.632                                \\
3                              & 90.19                                 & 78.92                                 & 20.57                                 & 0.674                                \\
4                              & { \textbf{92.68}} & { \textbf{79.43}} & { \textbf{21.94}} & { \textbf{0.68}} \\ \bottomrule
\end{tabular}}
\end{table}
\subsection{Ablation Study}
 We investigated the impact of the number of SimAM modules on the final segmentation and stitching outcomes. The quantitative results presented in Table \ref{table_Ablation} demonstrate that the embedded SimAM modules can significantly enhance segmentation performance, thereby improving the quality of stitching. Furthermore, as the number of SimAM modules increases, the quality of stitching also correspondingly improves.

\section{Discussion and conclusion}

In this paper, we propose an end-to-end two-stage X-ray medical image stitching method, where segmentation serves as a pre-task for stitching, aiming to reduce the complexity of searching for matching areas across the entire image. To this end, we designed the VMS-UNet to filter out the salient content of the image. For the stitching part, we designed a pedicle screw alignment energy function to guide the alignment, and finally used a hybrid energy function to estimate the optimal seam, thereby eliminating parallax artifacts. Experimental results show that our method outperforms  State of the Art (SOTA) stitching schemes on multiple key metrics.

\vfill\pagebreak

\bibliographystyle{IEEEbib}
\bibliography{strings,refs}

\end{document}